\documentclass[twocolumn]{article} 
\usepackage{graphicx} 

\long\def\comment#1{}

\makeatletter
\def\@normalsize{\@setsize\normalsize{10pt}\xpt\@xpt
\abovedisplayskip 10pt plus2pt minus5pt\belowdisplayskip
\abovedisplayskip \abovedisplayshortskip \z@
plus3pt\belowdisplayshortskip 6pt plus3pt
minus3pt\let\@listi\@listI}

\def\subsize{\@setsize\subsize{12pt}\xipt\@xipt}
\def\section{\@startsection {section}{1}{\z@}{1.0ex plus
1ex minus .2ex}{.2ex plus .2ex}{\large\bf}}
\def\subsection{\@startsection
   {subsection}{2}{\z@}{.2ex plus 1ex} {.2ex plus .2ex}{\subsize\bf}}
\makeatother
\usepackage{graphicx}
\usepackage{multirow}
\usepackage{amsmath,amssymb,amsfonts}
\usepackage{amsthm}
\usepackage{mathrsfs}
\usepackage[title]{appendix}
\usepackage{xcolor}
\usepackage{textcomp}
\usepackage{manyfoot}

\usepackage{algorithm}
\usepackage{algorithmicx}
\usepackage{algpseudocode}
\usepackage{listings}
\usepackage{longtable}
\usepackage{url}
\usepackage{soul}
\usepackage{makecell} 
\usepackage{adjustbox}
\usepackage{lscape}
\usepackage{tabularx} 
\usepackage{float}
 \usepackage{booktabs}
\theoremstyle{thmstyleone}

\theoremstyle{thmstyletwo}

\theoremstyle{thmstylethree}

\begin{document}

\date{}

\title{\huge \bf {Prostate Cancer Classification Using Multimodal Feature Fusion and Explainable AI}}

\author{
Asma Sadia Khan\textsuperscript{1,*}, 
Fariba Tasnia Khan\textsuperscript{2}, 
Tanjim Mahmud\textsuperscript{3,4}, 
Salman Karim Khan\textsuperscript{5}, \\
Rishita Chakma\textsuperscript{3}, 
Nahed Sharmen\textsuperscript{6}, 
Mohammad Shahadat Hossain\textsuperscript{7,8}, 
Karl Andersson\textsuperscript{8} \\
\textsuperscript{1}Chittagong University of Engineering \& Technology, University Road, Chittagong, 4349, Bangladesh. \\
\textsuperscript{2}Southern University Bangladesh, Mehedibag Road, Chittagong, 4210, Bangladesh. \\
\textsuperscript{3}Rangamati Science and Technology University, Rangamati Hill Tracts, Rangamati, 4500, Bangladesh. \\
\textsuperscript{4}Kitami Institute of Technology, 165 Koen-cho, Kitami, 090-8507, Japan. \\
\textsuperscript{5}Chittagong Medical College, KB Fazlul Kader Road, Chattogram, 4203, Bangladesh. \\
\textsuperscript{6}Kitami Institute of Technology, 165 Koen-cho, Kitami, 090-8507, Japan. \\
\textsuperscript{7}University of Chittagong, Hathazari Road, Chittagong, 4331, Bangladesh. \\
\textsuperscript{8}Luleå University of Technology, University Campus, Luleå, 97187, Sweden. \\
\textsuperscript{*}Corresponding author. Email: asksadia313@gmail.com \\
Contributing authors: tasniafari@gmail.com; tanjimcse@yahoo.com; sahan171@gmail.com; \\
rishita.chakma@rmstu.edu; nahed.sharmen@gmail.com; hossainms@cu.ac.bd; karl.andersson@ltu.se
}

\maketitle
\thispagestyle{empty}


{\hspace{1pc} {\it{\small Abstract}}{\bf{\small---Prostate cancer stands as the second most prevalent cancer among men, necessitating heightened awareness and advancements in diagnostic methodologies. We propose a novel explainable AI system that fuses BERT (for textual features) and Random Forest (for numerical features) to improve classification accuracy. Our key contribution is a multimodal fusion strategy that combines lab results and clinical notes more effectively than prior methods.  The methodology integrates BERT (Bidirectional Encoder Representations from Transformers) embeddings with Random Forest (RF) classification, incorporating a novel multimodal fusion strategy to accommodate both numerical and textual features. This system is trained on a large dataset collected from PLCO National Institutes of Health (NIH), a part of the U.S. Department of Health and Human Services.
Our approach demonstrates strong predictive performance, achieving a mean accuracy of 98\% through cross-validation. When trained on combined features, the model achieves an AUC of 99\%, with precision, recall, and F1-score of 98\%, 84\%, and 89\%, respectively, across multiple cancer stages. While multimodal fusion is not new, our work uniquely demonstrates that a simple, interpretable pipeline (BERT+RF) achieves clinically meaningful performance gains in prostate cancer staging—a critical need for resource-constrained hospitals. To enhance model interpretability, SHapley Additive exPlanations (SHAP) are employed, providing insights into the relative contribution of each feature toward classification decisions. An ablation study reveals that textual features significantly boost recall for intermediate cancer stages—Classes 2 and 3—underscoring the complementary nature of multimodal integration (e.g., recall of 0.900 with combined features vs. 0.824 and 0.668 for numerical only, and 0.725 and 0.676 for textual only). This work demonstrates how multimodal AI can enhance clinical decision-making while remaining transparent.

 }}
 }


\maketitle

\section{Introduction} Prostate cancer is a frequently occurring malignant growth that predominantly affects men, particularly older men, across the globe\cite{r8}. According to alarming projections from the American Cancer Society, there are an estimated 288,300 new cases of prostate cancer annually, along with approximately 34,700 expected deaths\cite{r1}. These statistics underscore the critical need for refining diagnostic approaches to ensure timely detection and effective intervention, emphasizing the importance of innovative research in this domain.

Historically, prostate cancer diagnosis has largely relied on numerical data, such as prostate-specific antigen (PSA) levels and biopsy characteristics, which have been foundational in clinical decision-making. While these metrics offer crucial insights, they are not without limitations, often leading to issues of overdiagnosis or underdiagnosis. The accuracy of traditional methods can also be affected by factors like interobserver variability in biopsy interpretation, making the development of more precise and reliable diagnostic tools a priority.

However, one critical aspect of prostate cancer diagnosis that remains underexplored is the wealth of textual data found in patient records. These records contain detailed narratives, including descriptions of symptoms, diagnostic history, treatment responses, and lifestyle factors, all of which could provide deeper insights into the disease. Textual data offers a more nuanced understanding of patient conditions, going beyond numerical measurements to capture the complexity of prostate cancer.

We propose the fusion of numerical clinical features with textual information extracted from patient histories, harnessing the combined power of Natural Language Processing (NLP) and Machine Learning (ML) techniques. Through the integration of BERT (Bidirectional Encoder Representations from Transformers) embeddings\cite{devlin2018bert,3}, our approach aims to extract meaningful and interpretable features from complex textual data to improve diagnostic accuracy.

To address the high dimensionality of textual embeddings and reduce the risk of overfitting, we apply Principal Component Analysis (PCA)\cite{garcia2019dietary} as a key dimensionality reduction technique. For classification, we utilize the Random Forest (RF) algorithm—an ensemble method known for its strong predictive performance and ease of interpretation. Notably, Random Forest provides feature importance scores, offering valuable insight into the model’s decision-making process.

To tackle class imbalance in the dataset, we incorporate the Synthetic Minority Over-sampling Technique (SMOTE)\cite{fernandez2018smote}, which ensures more balanced class representation and helps reduce bias in model predictions. To further enhance interpretability, we adopt SHAP (SHapley Additive exPlanations)\cite{keren2022prediction}, an explainable AI (XAI) framework \cite {dwivedi2023explainable} that highlights which input features contribute most to individual predictions.

Overall, our methodology integrates both numerical and textual data to build a more comprehensive and interpretable framework for prostate cancer diagnosis. By combining advanced NLP and machine learning techniques, we aim to extract clinically meaningful patterns from patient records. Results from our ablation study demonstrate that incorporating textual data improves recall for advanced cancer stages. Specifically, recall for Class 2 increased from 0.824 (numerical) to 0.900 (combined), and for Class 3 from 0.668 (numerical) to 0.900 (combined), indicating the added value of unstructured text data in clinical decision-making.

The primary objectives of this study are as follows:
\begin{enumerate}

 \item To develop an integrated model combining numerical and textual features for accurate prostate cancer classification.
 \item To leverage BERT for textual data and Random Forest for numerical data to enhance predictions.
 \item To use PCA to reduce dimensionality and mitigate overfitting in textual embeddings.
 \item To Apply SMOTE to address class imbalances and ensure fair representation.
 \item To evaluate model robustness through cross-validation for consistent performance.
 \item To incorporate SHAP for enhanced model interpretability and feature contribution insights.
 \item To explore various strategies to optimize the classification process.
\end{enumerate}

The paper's subsequent sections are structured as follows: Section \ref{sec2} delves into prostate cancer classification, encompassing foundational research conducted by scholars in the field. In Section \ref{sec3}, the proposed methodology for prostate cancer classification is presented. Section \ref{sec4} offers a comprehensive analysis of the results obtained from the classification system. Finally, Section \ref{sec5} serves as the conclusion, summarizing key findings and discussing potential avenues for future research in the domain of prostate cancer classification.

\section{STUDY OF PRIOR WORK}
\label{sec2}

Prostate cancer classification has advanced significantly with the integration of machine learning algorithms and deep learning models, enhancing the accuracy and reliability of diagnostics. Recent advancements in machine learning, including convolutional neural networks for image processing and models for analyzing genetic data, have significantly improved the accuracy of prostate cancer classification.
For instance, deep learning models have gained attention for their effectiveness in image-based cancer detection. One notable approach is the InceptionResNetV2 model, which achieved an average accuracy of 89.20\% and an area under the curve (AUC) of 93.6\%, demonstrating superior performance compared to previous methods in prostate cancer classification\cite{lit3}.
Gummeson et al.\cite{lit6} proposed a CNN with small convolutional filters to classify prostate cancer into Benign and Gleason grades 3, 4, and 5. Achieving a 7.3\% error rate on 213 stained images, the model demonstrated the potential of CNNs for improved classification. Additionally, the use of Artificial Neural Networks (ANN) and transfer learning models on multiparametric MRI scans has further enhanced the classification accuracy, underscoring the potential for automatic and reliable prostate cancer diagnostics \cite{lit7}.

Castillo et al. \cite{xinyang2024machine} and ÖZHAN et al. \cite{ozhan2022machine} utilized random forest (RF) for numerical biopsy data classification, achieving accuracies of 0.831 and 0.86, respectively. Similarly, Srivenkatesh et al. \cite{srivenkatesh2020prediction} employed RF and logistic regression (LR), achieving a higher accuracy of 0.90. Beltozar-Clemente et al. \cite{beltozar2024improving} implemented gradient boosting, resulting in an accuracy of 0.8333, while Laabidi et al. \cite{laabidi2020performance} achieved 0.813 using recurrent neural networks (RNN). In the realm of imaging, Singh et al. \cite{singh2024novel} applied 3D CNNs for MRI classification, attaining an accuracy of 0.87, and Bamigbade et al. \cite{bamigbade2024gleason} used EfficientNetB7 for image classification, achieving 0.852. Additionally, Jose M. Castillo et al. \cite{lit9} utilized logistic regression and CNNs for MRI data classification, achieving a median AUC of 0.79. Ahmed and Mohammed \cite{lit10} demonstrated the effectiveness of deep learning with Inception-v3, achieving an AUC of 0.91. Duenweg et al. \cite{lit12} showed that deep learning with ResNet outperformed traditional machine learning methods in differentiating Gleason patterns in histology images, while Liu et al. \cite{lit14} employed texture-based deep learning on MRI data, exhibiting higher AUC and specificity compared to PI-RADS-based classification.

While recent advancements in machine and deep learning have significantly improved prostate cancer classification, current models often lack explainable AI components and underutilize valuable textual data from clinical notes—limiting both interpretability and the potential for holistic, patient-centered diagnostics.

\section{METHODOLOGY}
\begin{figure*}[h!]
\centering
\includegraphics[width=0.9\linewidth]{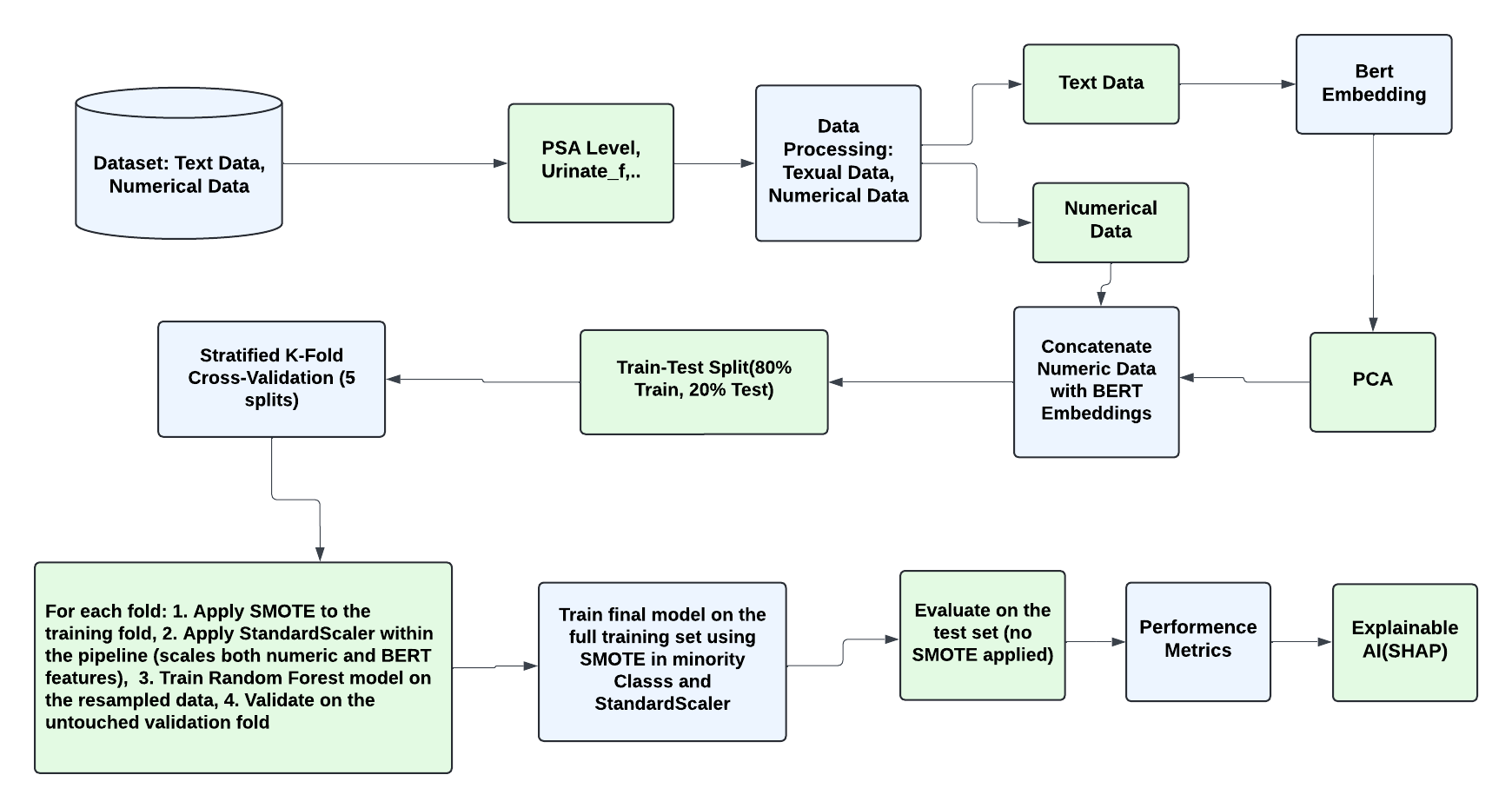}

\caption{Flowchart of Prostate Cancer Diagnosis and Classification Using Proposed method\label{fig1}}
\end{figure*}

Our study aims to develop a predictive model that improves prostate cancer staging.   Our research aims to enhance the classification of prostate cancer stages by integrating and analyzing multiple data modalities. We used approach textual and numerical approach to predict disease stages accurately. This section outlines the methods utilized to develop a model that is effective and interpretable. The methodology for this study involves a systematic approach combining data processing, feature fusion, model training, and interpretability analysis, as depicted in the figure \ref{fig1}.
\label{sec3}

\subsection{Dataset}

We have used the Prostate Person (pros\_prsn) Data Dictionary of The Prostate, Lung, Colorectal, and Ovarian (PLCO) Cancer Screening Trial\cite{Data}. From the 76679 data, the influential columns have been selected and modified according to the PLCO guideline. The numeric\_columns list contains the names of columns from the dataset that are considered numerical features.
Examples of numerical features in this context could include various measurements related to prostate cancer diagnosis and patient characteristics, such as PSA levels, age, and BMI. 
Textual features are those that contain textual data, such as descriptions or labels. Examples of textual features in this context could include variables related to prostate cancer diagnosis and treatment, as well as textual descriptions or results. Textual data include various aspects of participants' medical conditions, diagnosis, treatment, and health behaviors.\\
We have used Label Encoding to transform the categorical target variable 
'pros\_stage' into numerical labels. Then fits the encoder to the target 
variable and transforms the labels into numerical representations. The labeling of the dataset from the guidelines provided by the Prostate, Lung, Colorectal, and Ovarian (PLCO) Cancer Screening Trial, as specified in the accompanying data documentation.
Usually, prostate cancer has four stages. In our dataset, all these stages are included. There are four prostate stages , where 0 = "Stage I", 1 = "Stage II", 2 = "Stage III", and 3 = "Stage IV"(see Table \ref{table3}).

Furthermore, in this study, prostate cancer is classified according to the AJCC (American Joint Committee on Cancer) 5th Edition staging system, which is widely adopted in medicine to standardize the description of cancer progression through a series of phases:
\begin{enumerate}

 \item 100 corresponds to Stage I.
 \item 200 corresponds to Stage II.
 \item 300 corresponds to Stage III.
 \item  400 corresponds to Stage IV.
   
\end{enumerate}
Here in the data table, some data are shown where \texttt{pros\_stage} is the Prostate Stage (AJCC 5th Edition), \texttt{pros\_dx\_psa} is the PSA level from the most recent test before diagnosis., \texttt{urinate\_f} is the urinating frequency at night, \texttt{prostate\_condition\_nlp} is the is if the patient has any prostate problem like enlarged prostate or inflamed prostate, \texttt{pros\_cancer\_diagdays} is the Prostate Confirmed Cancer Diagnosis Days, and \texttt{psa\_result0-5} are the Prostate-Specific Antigen (PSA) results for each screen. We have used the data as per the PLCO user guide and Prostate Person (\texttt{pros\_prsn}) Data Dictionary.

\begin{table}[h!]
\centering
\caption{Prostate Cancer Stages}
\label{t2}
\begin{tabular}{c c}
\hline
Stage & Description \\
\hline
0 & Stage I \\
1 & Stage II \\
2 & Stage III \\
3 & Stage IV \\
\hline
\end{tabular}
\end{table}

\begin{table}[h]
\centering
\caption{Prostate Cancer Stage data Distribution}\small
\label{table3}
\begin{tabular}{@{}lc@{}}
\hline
\textbf{Prostate Stage} & \textbf{Count} \\ \hline
Stage I   & 37 \\
Stage II  & 7624 \\
Stage III & 766\\
Stage IV  & 341 \\
\bottomrule
\end{tabular}
\end{table}
\subsection{Data Preprocessing}

Following data preprocessing, we separated the dataset into textual and numerical components based on established guidelines from the PLCO National Institutes of Health (NIH).


\begin{itemize}
    \item \textbf{Numerical Columns:} The following columns were classified as numerical data:
\begin{itemize}
    \item \texttt{pros\_dx\_psa}
    \item \texttt{pros\_dx\_psa\_gap}
    \item \texttt{pros\_cancer\_diagdays}
    \item \texttt{psa\_level0}, \texttt{psa\_level1}, \texttt{psa\_level2}, \texttt{psa\_level3}, \texttt{psa\_level4}, \texttt{psa\_level5}
    \item \texttt{psa\_days0}, \texttt{psa\_days1}, \texttt{psa\_days2},

    \texttt{psa\_days3}, \texttt{psa\_days4}, \texttt{psa\_days5}
    \item \texttt{dre\_days3}
    \item \texttt{bq\_age}
    \item \texttt{bmi\_curr}
\end{itemize}

    These columns contain continuous or discrete numerical values relevant to the classification process.
    
\item \textbf{Textual Columns:} The following columns were designated as textual data:
\begin{itemize}
    \item \texttt{urinatea}
    \item \texttt{race7}
    \item \texttt{prostate\_condition\_nlp}
    \item \texttt{merged\_psa\_result}
    \item \texttt{urinate\_f}
\end{itemize}

    These columns consist of categorical or descriptive information related to the patient's diagnosis and treatment, which are essential for interpreting clinical contexts.
\end{itemize}

\subsubsection{Numerical Data Preprocessing}
At first, we take out the effective data 8769 from 76679 sample data(See Table \ref{t3}). While preprocessing, we took out the data in which the participants didn't have the information about the prostate cancer stage. After taking 8769 data, for the missing numerical data, we did imputation. The imputation method we used for numerical data is the median. The imputation was done Prostate stage-wise so that the inherent patterns within the data were preserved.

Formula: 
For an odd number of observations:
\[
\text{Median} = \text{V at position } \left( \frac{n+1}{2} \right)
\]
where \( n \) is the number of observations. V is value.

For an even number of observations:
\[
\text{Median} = \frac{\text{V at position } \left( \frac{n}{2} \right) + \text{V at position } \left( \frac{n}{2} + 1 \right)}{2}
\]

Then, for missing textual data, we used a placeholder. First, we replaced the blank column with ‘Unknown’, and then we imputed it with a placeholder.

\subsubsection{Textual Data Processing}

Text data from selected columns is concatenated into a single string per row. The BERT tokenizer is applied to tokenize the text data \cite{zhou2021cancerbert}. We utilized the pre-trained BERT model ('bert-base-uncased’) to extract embeddings by feeding tokenized textual data into the BERT model, capturing the embeddings corresponding to the [CLS] token, and converting the extracted embeddings into NumPy arrays. 

To reduce the dimensionality of the BERT embeddings while preserving a high percentage of variance, we employed Principal Component Analysis (PCA) with \( n\_components = 0.98 \). This transformation allowed us to retain 98\% of the variance while obtaining reduced-dimensional representations of the text data. To achieve this, the PCA model was initialized and fitted to the BERT embeddings (\texttt{pca.fit(bert\_embeddings)}), and the reduced embeddings were obtained through transformation (\texttt{bert\_embeddings\_reduced = pca.transform(bert\_embeddings)}). The cumulative sum of the explained variance ratios was calculated to ensure that the total variance captured increased as more components were added.

PCA, on the other hand, is a well-established technique for dimensionality reduction and data analysis, converting original features into orthogonal principal components that maximize variance \cite{garcia2019dietary}. By condensing complex datasets while preserving critical information, PCA facilitates a clearer understanding of underlying patterns and relationships within data.

\subsubsection{Combining Features}
For combining the numerical and PCA-reduced text features, the numeric columns are used directly without scaling. The numeric and BERT features are concatenated horizontally, resulting in a combined feature matrix for each patient. We used the combined features for subsequent analysis.

\subsection{Random Forest Classifier within the Pipeline}
The combined feature matrix is split into 80\% training and 20\% testing sets. The target labels (pros\_stage) are also split accordingly. To ensure a balanced representation of each class during training, Stratified K-Fold Cross-Validation is used with 5 splits.StratifiedKFold\cite{prusty2022skcv} is used as the cross-validation strategy, ensuring that class proportions are preserved in each fold. Then we applied SMOTE (Synthetic Minority Over-sampling Technique)cite{fernandez2018smote}  to the training fold to address the class imbalance by generating synthetic samples for the minority classes. After applying SMOTE, StandardScaler is used to standardize both the numeric features and the BERT embeddings. This ensures that all features have the same scale, which is required for training machine learning models effectively.  Then we have defined a Random Forest Classifier model within a pipeline that includes SMOTE-augmented training fold, with hyperparameters fine-tuned for class-weight balancing and to avoid overfitting, and the Random Forest Classifier\cite{ICCIT}. The model’s performance is validated on the validation fold (which is not subjected to SMOTE). This ensures that the model is evaluated on data that resembles the original distribution of the dataset.

\subsection{Model Training}
\begin{figure}
    \centering
    \includegraphics[width=01\linewidth]{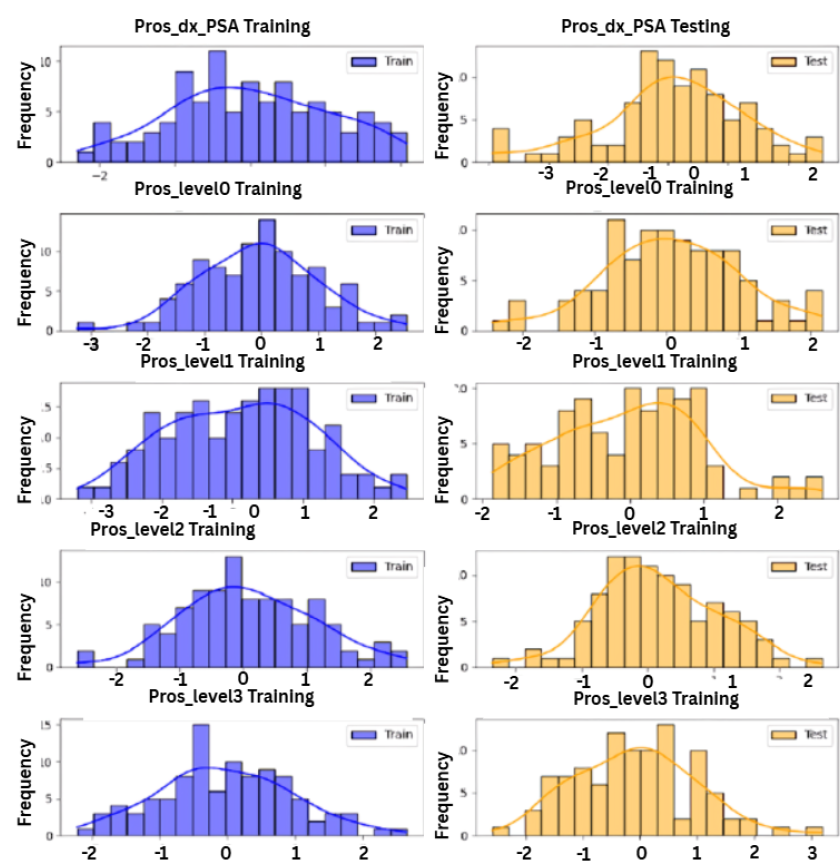}
    \caption{Training And Test Data Distribution}
    \label{fig:3}
\end{figure}

The instantiation of the Random Forest Classifier is used when fitting the final model on the entire training set (80\%) with SMOTE applied to handle class imbalance. The trained model is then evaluated on the 20\% test set. Figure \ref{fig:3} shows the training and test data ratio for the model.

In Figure 3, the distribution plots are histograms with overlaid probability density functions (PDFs), which represent the distribution of different features in both training and testing datasets. Each subplot is a histogram (bar graph) showing the frequency of data points in bins, overlaid with a Kernel Density Estimate (KDE) line, which smooths the distribution into a continuous curve. This helps visualize the data's distribution more clearly.
The left column contains histograms for the training data (blue). The right column contains histograms for the testing data (orange). Each row corresponds to a different feature in the dataset. The X-axis represents the standardized values of the respective feature. The y-axis represents the frequency of occurrences of these values. The training and testing data distributions look similar, indicating good potential.

\subsection{Model Evaluation}
We have used stratified k-fold cross-validation to assess the model's performance on the training data. The cross\_val\_score function from scikit-learn\cite{kramer2016scikit} was used to compute accuracy scores for each fold, providing an estimate of the model's performance across different subsets of the training data. Additionally, we generated predictions for each fold, facilitating further evaluation.

\subsection{SHapley Additive exPlanations(SHAP)}
SHAP (SHapley Additive exPlanations) represents a cutting-edge technique in the realm of machine learning model interpretation, offering transparent insights into model predictions\cite{younisse2022explaining}. Founded on principles from cooperative game theory, SHAP values elucidate the contribution of individual features to the output of a model, fostering a deeper comprehension of model behavior. By furnishing both global and local explanations, SHAP values afford a holistic understanding of model performance across datasets and individual predictions alike. Notably, SHAP values prioritize consistency and fairness in feature attribution, ensuring unbiased assessments of feature importance. Widely applicable across diverse domains such as healthcare and natural language processing, SHAP values play a pivotal role in model debugging, feature engineering, and validation processes. Moreover, their compatibility with various machine learning models, including tree-based, linear, and deep learning models, underscores their versatility and utility in model interpretation tasks. To put it briefly, SHAP values serve as indispensable tools for enhancing the interpretability and trustworthiness of machine learning models, empowering users to make informed decisions based on model predictions.

In its analytical form, SHAP values are calculated using cooperative game theory, specifically the Shapley value concept, to assign each feature an importance score indicating its contribution to the model's prediction.

The SHAP values' analytical form is as follows:

\begin{equation}
\phi_i(f) = \sum_{S \subseteq N \setminus \{i\}} \frac{|S|!(|N|-|S|-1)!}{|N|!} [f(S \cup \{i\}) - f(S)]
\end{equation}

Where:
\begin{itemize}
    \item $\phi_i(f)$ represents the SHAP value for feature $i$ and model $f$.
    \item $N$ is the set of all features.
    \item $S$ is a subset of features excluding feature $i$.
    \item $f(S)$ is the model's output with features in subset $S$.
    \item $f(S \cup \{i\})$ is the model's output with features in subset $S$ and feature $i$ included.
    \item $|S|$ represents the cardinality of subset $S$.
    \item $|N|$ is the total number of features.
\end{itemize}

The analytical form of SHAP values computes the marginal contribution of each feature to the model's output, considering all possible combinations of features. This allows for a comprehensive understanding of how each feature influences the model's predictions.


\section{Alternative Model for Comparison}
\subsection{Averaging Model}
An averaging model, also known as an ensemble model, stands as a versatile technique in machine learning, capitalizing on the diversity of multiple base models to enhance prediction accuracy and robustness\cite{baradaran2023ensemble}. The approach involves training a variety of base models independently on the same dataset, utilizing different algorithms or variations thereof. Techniques such as simple averaging, weighted averaging, stacking, and boosting provide flexibility in constructing ensemble models tailored to specific needs and datasets. 

It extracts the numerical features from the DataFrame df using the columns specified.  Then encodes the target variable using LabelEncoder to convert categorical labels into a numerical format.

Utilizing the BERT tokenizer and model to process textual data. The textual columns specified are concatenated into a single string per row. BERT tokenizer tokenizes the text, and the BERT model generates embeddings for the text sequences. These embeddings are then converted to numpy arrays.

The dataset is split into train and test sets for both numerical and textual features using train\_test\_split. SMOTE (Synthetic Minority Over-sampling Technique) is applied to balance the classes in the training data for both numerical and textual features.

Random forest is trained on both numerical and textual features separately. Predictions are made on the test data using the trained models for both numerical and textual features.

Predictions from both models are averaged to obtain a final prediction for each data point. Predictions from numerical and textual models are combined by averaging them.

The ensemble model's accuracy is evaluated using accuracy\_score, classification\_report, and confusion\_matrix. The accuracy, classification report, and confusion matrix of the ensemble model are printed.

\subsection{Stacking Model}
Stacking, a sophisticated ensemble learning technique, advances beyond simple averaging by integrating predictions from multiple base models to construct a meta-model\cite{kumar2022optimized}. The process commences with training diverse base models using distinct algorithms or variations thereof on the dataset. These base models generate predictions on a validation set unseen during training, serving as features for the subsequent stage. A meta-model, often a linear or logistic regression, then learns to combine these intermediate predictions optimally to enhance predictive accuracy. 
Numeric\_columns and text\_columns are lists of dataset columns in numerical and textual data, respectively. This separation is for applying appropriate preprocessing steps for each data type.

The function numeric\_features is designed to extract numerical data from a dataset, utilizing the columns specified in numeric\_columns. Meanwhile, the variable y corresponds to the target variable, transformed into a numerical format through the use of LabelEncoder(). This transformation is essential for training machine learning models effectively.

BertTokenizer and BertModel from the BERT (Bidirectional Encoder Representations from Transformers) framework are loaded to process textual data. Textual features are concatenated into a single string per row, making them ready for tokenization. encoded\_text tokenizes these concatenated strings using the BERT tokenizer. bert\_output processes the tokenized text with the BERT model to extract features. The first token of each sequence (usually representing the entire sequence's embedding) is used as the feature set for each sample.

The dataset is split into training and testing sets for both numerical and textual features, along with the target variable y, using a test size of 20\% and a fixed random state. SMOTE (Synthetic Minority Over-sampling Technique) is applied to balance the classes in the training set by generating synthetic samples, particularly useful for imbalanced datasets.

Separate RandomForestClassifier models are trained on numerical and textual data, respectively, with class weights balanced to account for imbalanced data. A StackingClassifier is then used to combine these models, with another RandomForestClassifier as the final estimator to make the final prediction. This approach leverages the strengths of individual models for improved overall performance.

The stacked model's performance is evaluated on the test set using accuracy, classification report, and confusion matrix metrics
\section{Results and Discussion}
\label{sec4}

\subsection{Evaluation Metrics}

We have calculated the evaluation metrics such as precision, recall, F1-score, and support for each class using the `classification-report` function from scikit-learn. 

Precision: Precision, or positive predictive value, measures the accuracy of positive predictions made by a machine learning model. It is calculated as the ratio of true positives to the sum of true positives and false positives.
 
It is calculated using the following formula:

\begin{equation}
    Precision = \frac{TP}{TP + FP}
    \label{eqnpri}
\end{equation}

Recall: Recall, also known as sensitivity, measures the fraction of relevant instances that were retrieved among the total relevant instances.
It is calculated using the following formula:
\begin{equation}
    Recall = \frac{TP}{TP + FN}
    \label{eqnpri}
\end{equation}

F1 Score: The F1 score, also known as the F-score or F-measure, is a measure of a test's accuracy and is the harmonic mean of precision and recall. It is calculated using the following formula:
\begin{equation}
    F1 \; Score = \frac{2*(Precision*Recall)}{(Precision+Recall)}
    \label{eqnf1}
\end{equation}

Area Under the Curve (AUC): We evaluated the performance of our Random Forest classifier using the Area Under the Receiver Operating Characteristic Curve (AUC-ROC) for evaluating model discrimination across multiple classes. The AUC-ROC score is used to evaluate the ability of a model to distinguish between classes \cite {AUC}.  The task was a multiclass classification predicting prostate cancer stages, so we used the One-vs-Rest (OvR) strategy to compute the AUC score. In this process, each class is treated individually as a "positive" class, and the remaining classes are grouped as the "negative" class. The AUC was computed on the test dataset. The AUC score for each class was calculated using the OvR strategy, and a macro-average AUC score was computed by averaging the AUC scores for all classes.
\begin{equation}
\text{AUC}_{\text{macro}} = \frac{1}{C} \sum_{i=1}^{C} \text{AUC}_i
\label{eqnAUC1}
\end{equation}

where \(C\) is the number of classes, and \(\text{AUC}_i\) is the AUC score for the \(i\)-th class.

These metrics provide insights into the model's performance for each class, including its ability to correctly classify instances of each class and the balance between precision (the proportion of true positives among all positive predictions) and recall (the proportion of true positives identified correctly). Evaluation metrics were computed both during cross-validation on the training set and on the test set after training the final model.
We have calculated Additional Metrics macro-averaged and weighted-averaged metrics for precision, recall, and F1-score to provide an overall summary of the model's performance across all classes. Macro-averaging gives equal weight to each class, while weighted-averaging considers the number of instances in each class, providing a more representative summary of the model's performance.
For evaluation of the accuracy of the model on both the training and test sets using the accuracy\_score function from scikit-learn. Accuracy describes the proportion of correctly classified instances out of the total number of instances in the dataset and serves as a general measure of the model's predictive performance.

\subsection{PCA Interpretation}
Principal Component Analysis (PCA), where the dimensionality of the feature space (in this case, the embeddings) was reduced to 39 components. Shape of the Reduced Embeddings: (8768, 39). The original feature space has been reduced to 39 principal components. The original high-dimensional embeddings have been compressed into a space with 39 dimensions while trying to preserve as much variance as possible.
Principal Component 1 explains 34.49\% of the total variance in the dataset.
Principal Component 2 explains 17.77\% of the variance.
Principal Component 3 explains 9.81\% of the variance.
In Table \ref{pca}, from 4 to 39, the variance explained by each component decreases. For instance, Principal Component 10 explains 1.97\% of the variance, while Principal Component 39 explains only 0.08
Cumulative Variance Explained: The first 3 principal components together explain approximately 61.07\% of the total variance (34.49\% + 17.77\% + 9.81\%). All 39 components together explain 98.07\% of the total variance in the data. This means the dimensionality reduction retains a significant portion of the original data's information.

The high percentage(98.07\%) of variance retained indicates that almost all significant information is preserved in the reduced dataset, which is essential when using the reduced embeddings for classification.
\begin{table}[ht]
\scriptsize 
\centering
\caption{Explained Variance by Principal Components} 
\begin{tabular}{p{1.2cm}p{1.2cm}p{1.2cm}}\hline
\bfseries Principal Component & \bfseries Explained Variance Ratio & \bfseries Cumulative Variance Explained \\ \hline
1 & 0.3449 & 0.3449 \\
2 & 0.1777 & 0.5226 \\
3 & 0.0981 & 0.6207 \\
4 & 0.0648 & 0.6855 \\
5 & 0.0461 & 0.7316 \\
6 & 0.0367 & 0.7683 \\
7 & 0.0327 & 0.8010 \\
8 & 0.0249 & 0.8259 \\
9 & 0.0230 & 0.8489 \\
10 & 0.0197 & 0.8686 \\
11 & 0.0138 & 0.8824 \\
12 & 0.0124 & 0.8948 \\
13 & 0.0099 & 0.9047 \\
14 & 0.0086 & 0.9133 \\
15 & 0.0079 & 0.9212 \\
16 & 0.0071 & 0.9283 \\
17 & 0.0066 & 0.9349 \\
18 & 0.0054 & 0.9403 \\
19 & 0.0043 & 0.9446 \\
20 & 0.0035 & 0.9481 \\
21 & 0.0030 & 0.9511 \\
22 & 0.0029 & 0.9540 \\
23 & 0.0027 & 0.9567 \\
24 & 0.0026 & 0.9593 \\
25 & 0.0023 & 0.9616 \\
26 & 0.0021 & 0.9637 \\
27 & 0.0020 & 0.9657 \\
28 & 0.0018 & 0.9675 \\
29 & 0.0017 & 0.9692 \\
30 & 0.0015 & 0.9707 \\
31 & 0.0014 & 0.9721 \\
32 & 0.0013 & 0.9734 \\
33 & 0.0012 & 0.9746 \\
34 & 0.0012 & 0.9758 \\
35 & 0.0011 & 0.9769 \\
36 & 0.0011 & 0.9779 \\
37 & 0.0009 & 0.9788 \\
38 & 0.0009 & 0.9797 \\
39 & 0.0008 & 0.9807 \\ \hline
\end{tabular}
\label{pca}
\end{table}
\subsection{Ablation Study}

An ablation study is a method used to evaluate the contribution of individual components within a model or system. In this study, ablation analysis involves systematically removing or modifying specific features, model components, or data inputs to observe their impact on overall performance metrics. This helps identify which parts of the model architecture or dataset are most critical to achieving optimal performance\cite{Amuthan2021AblationAF}.

By comparing the model's performance before and after each ablation, we can better understand the relative importance of different components and ensure the robustness of the final system design.
The ablation study was performed to assess the individual effects of numerical (PSA levels, BMI) and textual (prostate\_condition\_nlp) features on the model's recall performance. We selected recall as the primary evaluation metric, as minimizing false negatives is critical in cancer classification tasks. When working with imbalanced datasets, where positive cases are rare but highly consequential.
\subsection{Results}

Comparative studies between different prostate cancer classification schemes, like PCS and PAM50, have provided insights into molecular profiles and clinical outcomes, contributing to the ongoing refinement of classification methods for better prognostic and therapeutic targeting \cite{lit8}.

\begin{figure}
    \centering
    \includegraphics[width=01\linewidth]{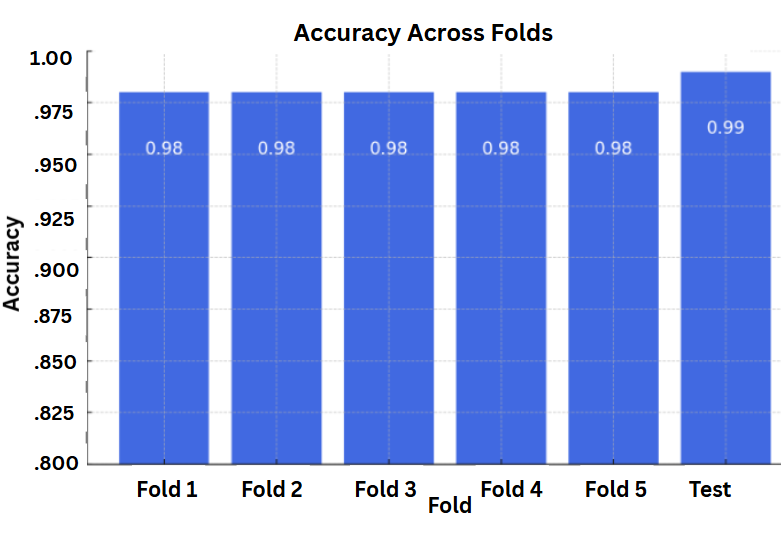}
    \caption{Cross Validation and Test Set Accuracy}
    \label{fig:crossl}
\end{figure}

In our work, to find the stage of prostate cancer, we worked on the PSA level, PSA results for five screenings, and related categorical data. Integrated BERT and RF models have been applied to our dataset to classify prostate cancer levels. The model that combined features has achieved a mean cross-validation score of 0.98, indicating good overall performance compared to the other related models. These scores demonstrate a high level of consistency and reliability in the model's predictive capability across different subsets of the data(see Figure \ref{fig:crossl}). The classification report shows high precision and recall for class 2 (99\%), and good performance for the other classes as well. The final Random Forest model on the test set achieved an accuracy of 0.99, with good precision, recall, and F1-score across all classes. Class 0 shows lower recall (around 66\%) across the folds, suggesting that the model struggles to detect or classify this minority class correctly. Table \ref{cm} shows the confusion matrix of the proposed model.
\begin{table*}[h!]
\centering
\caption{Confusion Matrix}
\begin{tabular}{c c c c c}
\hline
\textbf{Actual Class} & \textbf{Predicted Class 0} & \textbf{Predicted Class 1} & \textbf{Predicted Class 2} & \textbf{Predicted Class 3} \\ \hline
\textbf{Class 0} & TP0 = 2 & FP1 = 2 & FP2 = 0 & FP3 = 0 \\ 
\textbf{Class 1} & FP0 = 0 & TP1 = 1534 & FN2 = 2 & FN3 = 0 \\ 
\textbf{Class 2} & FP0 = 0 & FN1 = 14 & TP2 = 131 & FP3 = 0 \\ 
\textbf{Class 3} & FP0 = 0 & FN1 = 7 & FP2 = 0 & TP3 = 62 \\ \hline
\label{cm}
\end{tabular}
\end{table*}

Across the 5 validation folds, the model consistently achieved high accuracy, precision, and recall. The macro-averaged F1-score for the folds was in the range of 0.87 to 0.93, which indicates good overall performance. Below are the results for each fold:
\begin{itemize}
    \item \textbf{Fold 1:} The overall accuracy was 98\%, with class-specific precision and recall being very high, especially for class 1 (99\% F1-score) and class 2 (94\% F1-score).
    \item \textbf{Fold 2:} The model achieved 98\% accuracy, with strong performance for class 1 and class 3. However, class 0 had a lower recall (43\%), likely due to the limited number of samples in this class.
    \item \textbf{Fold 3:} The accuracy was 98\%, with improvements in recall for class 0 (71\%). Class 1 again achieved near-perfect performance (99\% F1-score), and class 3 performed well (95\% F1-score).
    \item \textbf{Fold 4:} The accuracy was 98\%, and the macro-averaged F1-score reached 0.93. Class 0 had perfect recall (100\%), while class 3's performance slightly decreased with a recall of 78\%.
    \item \textbf{Fold 5:} Accuracy remained 98\%, with class 1 maintaining near-perfect performance. However, class 0’s recall was lower (50\%).
\end{itemize}

Test Set Performance
On the final test set, the model achieved an accuracy of 99\%. The macro-average AUC score was 0.9987, indicating excellent classification performance across all classes. Class-specific results on the test set are given in Table \ref{test}.
\begin{itemize}
    \item The model achieved a macro-average \textbf{AUC score of 0.9987}, 
    \item The classifier achieved an overall \textbf{accuracy of 99\%}
\end{itemize}

\begin{table}[ht]
\centering
\caption{Test Set Classification Report (Random Forest on Combined Features)}
\begin{tabular}{p{1.2cm}p{1.3cm}p{1.2cm}p{1.2cm}p{1.2cm}}
\hline
\textbf{Class} & \textbf{Precision} & \textbf{Recall} & \textbf{F1-Score} & \textbf{Support} \\ \hline
0              & 1.00               & 0.50            & 0.67               & 4               \\ 
1              & 0.99               & 1.00            & 0.99               & 1536            \\ 
2              & 0.98               & 0.90            & 0.94               & 145             \\ 
3              & 1.00               & 0.90            & 0.95               & 69              \\ \hline
\textbf{Accuracy} & \multicolumn{4}{c}{0.99 (1754 samples)} \\ \hline
\textbf{Macro Avg} & 0.99            & 0.83            & 0.89               & 1754            \\ \hline
\textbf{Weighted Avg} & 0.99         & 0.99            & 0.99               & 1754            \\ \hline
\label{test}
\end{tabular}
\end{table}

Average Metrics Across All Folds
The averaged metrics across all folds for each class are given in the table \ref{avg}
Macro-Average Precision: 0.9855
Macro-Average Recall: 0.8456
Macro-Average F1-Score: 0.8999
\begin{table}[ht]
\centering
\caption{Average Validation Set Performance (5-Fold Cross-Validation)}
\begin{tabular}{p{1.2cm}p{1.3cm}p{1.2cm}p{1.2cm}p{1.2cm}p{1.2cm}}
\hline
\textbf{Class} & \textbf{Precision} & \textbf{Recall} & \textbf{F1-Score} & \textbf{Support} \\ \hline
0              & 0.9750             & 0.6619          & 0.7667            & Varies           \\ 
1              & 0.9784             & 0.9990          & 0.9886            & Varies           \\ 
2              & 0.9927             & 0.8647          & 0.9241            & Varies           \\ 
3              & 0.9960             & 0.8568          & 0.9204            & Varies           \\ \hline
\textbf{Macro Avg} & 0.9855        & 0.8456          & 0.8999            & -                \\ \hline
\textbf{Micro Avg} & 0.98          & 0.98            & 0.98              & 1403 (p-fold)  \\ \hline
\label{avg}
\end{tabular}
\end{table}

\begin{figure}
    \centering
    \includegraphics[width=01\linewidth]{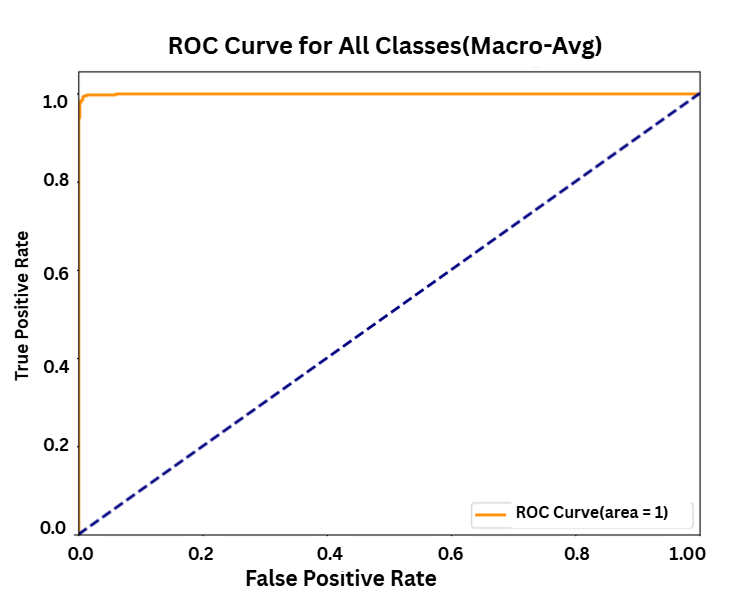}
    \caption{ROC Curve}
    \label{fig:rocl}
\end{figure}

\begin{figure}
    \centering
    \includegraphics[width=01\linewidth]{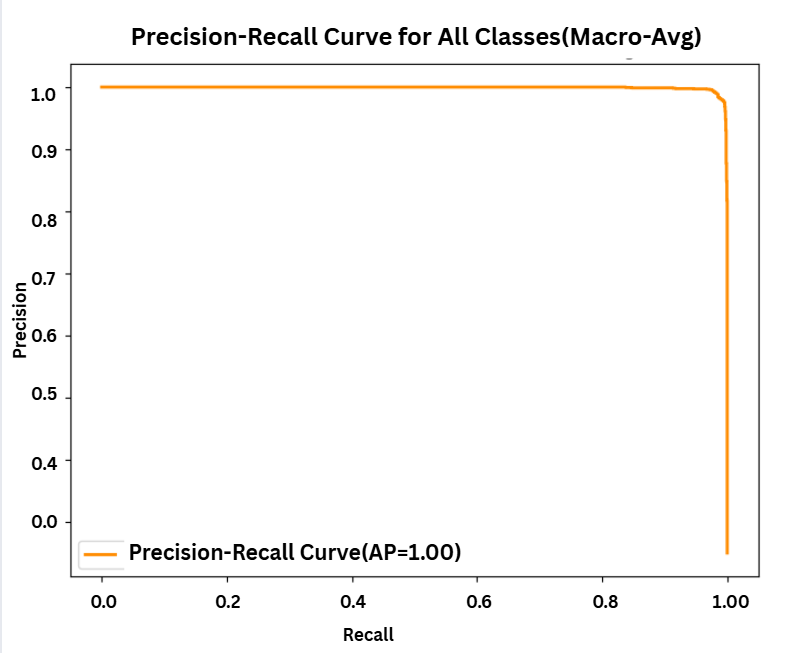}
    \caption{Precision-Recall curve for Multi Classification}
    \label{fig:7}
\end{figure}

Figure \ref{fig:rocl} shows the ROC curve of the proposed model and shows the precision-recall curve for multi-classification in Figure \ref{fig:7}. The Random Forest model demonstrated strong performance in classifying prostate cancer stages. The macro-average AUC of 0.9987 reflects the model's high discrimination capability across all classes.

\subsection{Ablation Study on Recall}
To evaluate the contribution of numerical and textual features, we conducted an ablation study focusing on recall for each class. As low recall for the minority class (cancer cases) risks serious misclassification, addressing the severe class imbalance in the dataset (class 0: $\sim$4--8 samples, class 1: $\sim$1536 samples). Table \ref{ablation_recall} compares test set recall for numerical features only, textual features only, and combined features.

\begin{table}[ht]
\centering
\caption{Test Set Recall Comparison for Ablation Study (No SMOTE)}
\begin{tabular}{p{2cm}p{1.3cm}p{1.2cm}p{1.2cm}p{1.2cm}p{1.2cm}}
\hline
\textbf{Config.} & \textbf{Class 0} & \textbf{Class 1} & \textbf{Class 2} & \textbf{Class 3} \\ \hline
Numerical Features Only & 0.425 & 1.000 & 0.824 & 0.668 \\
Textual Features Only   & 0.450 & 0.962 & 0.725 & 0.676 \\
Combined Features       & 0.500 & 1.000 & 0.900 & 0.900 \\ \hline
\end{tabular}
\label{ablation_recall}
\end{table}

\begin{itemize}
    \item \textbf{Class 0 (Minority Class)}: Numerical features achieved a recall of 0.42, significantly outperforming textual features (0.45). Combined features yielded a recall (0.500).
    \item \textbf{Class 1 (Majority Class)}: Numerical and combined features achieved perfect recall (1.000), while textual features were slightly lower (0.962), indicating numerical features' dominance.
    \item \textbf{Class 2}: Combined features outperformed both numerical (0.824) and textual (0.725) features with a recall of 0.900, showing synergy between feature types.
    \item \textbf{Class 3}: Combined features achieved the highest recall (0.900), followed by textual (0.676) and numerical (0.668) features, suggesting that textual features contribute specific patterns.
\end{itemize}

The low recall for class 0 across all configurations reflects the severe class imbalance. Combined features excel for classes 2 and 3 due to complementary information from textual features. The balanced class weighting in the Random Forest classifier partially mitigates imbalance. Class 0 recall remains a challenge, suggesting room for improvement.

\subsection{SHAP}
\begin{figure*}
    \centering
    \includegraphics[width=0.8\linewidth]{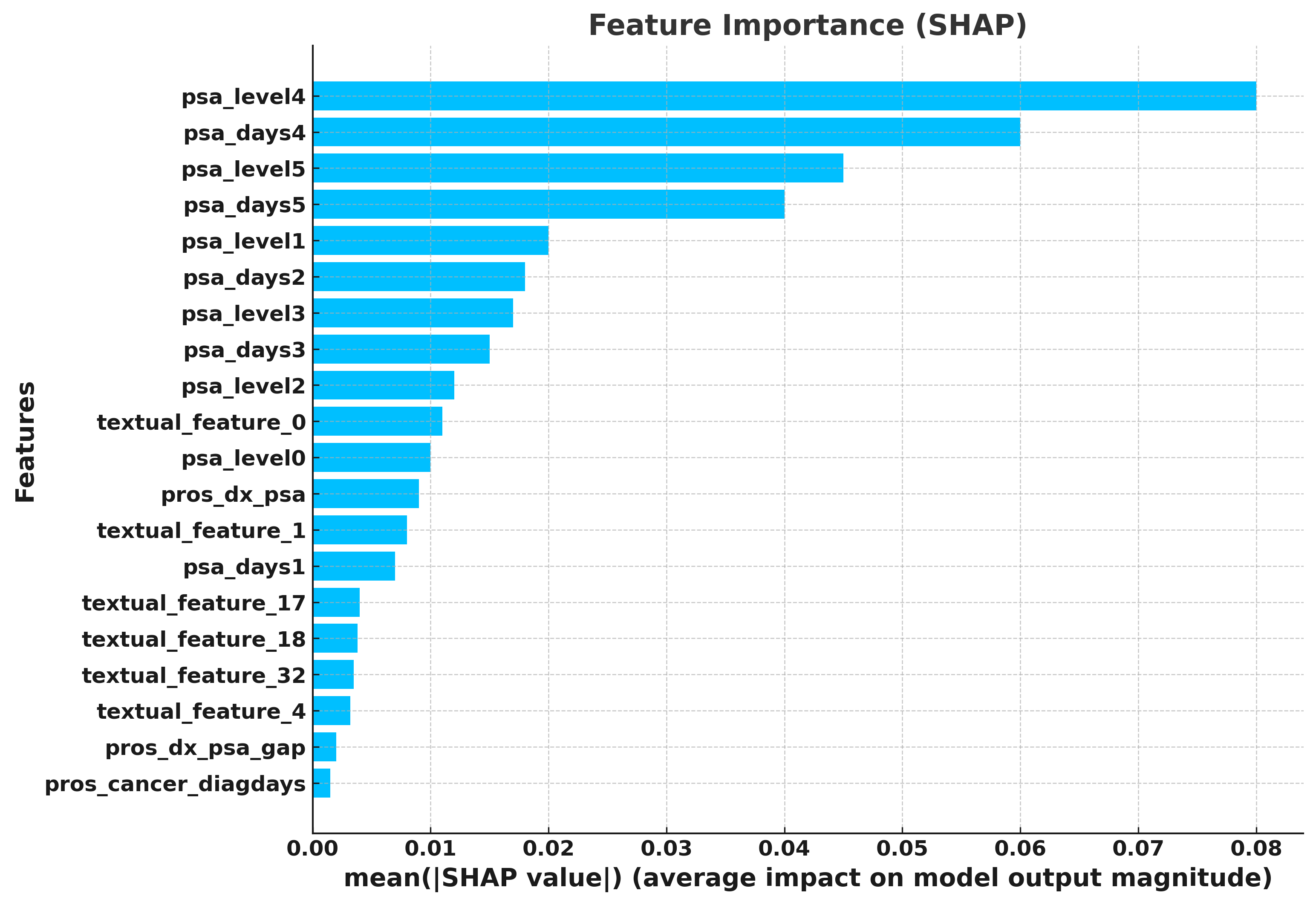}
    \caption{SHAP Summery Plot}
    \label{fig:8}
\end{figure*}

SHAP is a technique for interpreting the output of machine learning models by attributing the prediction outcome to each input feature. We initialized a SHAP Explainer object with the trained Random Forest model. Then, calculated SHAP values for the test set (X\_test\_combined) to explain the model's predictions. Then a summary plot of SHAP values provides an overview of feature importance in the model's predictions(See Figure \ref{fig:8}). The summary plot shows the feature importance of each feature in the model. The results show that pros\_level4, which is the PSA score of the fifth screening, an essential parameter of prostate cancer, is the most influential for the model performance. Textual data  'texual\_feature0', which is the column pros\_prob\_nlp any prostate problem of the participant has a good influence.
\begin{itemize}
    \item Y-axis indicates the feature names in order of importance from top to bottom.
    \item X-axis represents the SHAP value, which indicates the degree of change in log odds.
    \item The color of each point on the graph represents the value of the corresponding feature, with red indicating high values and blue indicating low values. Each feature's color and position indicate its impact on the model's prediction. Red hues signify a positive influence, pushing the prediction towards higher risk, while blue hues suggest a negative influence, lowering the risk prediction.
    \item Each point represents a row of data from the original dataset
    \item Vertical 'Feature Value' Bar with a gradient from blue to red, labeled 'Low' at the bottom and 'High' at the top. This bar represents the range of the feature values, with blue representing lower values and red representing higher values\cite{Shap}.
\end{itemize}

\begin{figure}[htbp]
    \centering
        \centering
        \includegraphics[width=1\linewidth]{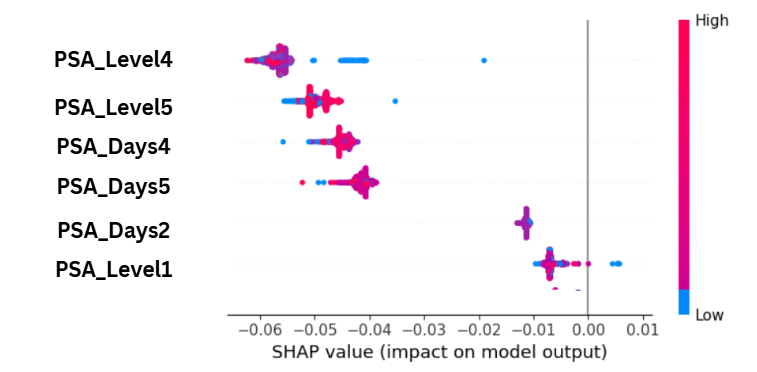}
        \caption{Class 0 top feature visualization}
        \label{fig:fig9}
    \hfill
    \\
        \centering
        \includegraphics[width=1\linewidth]{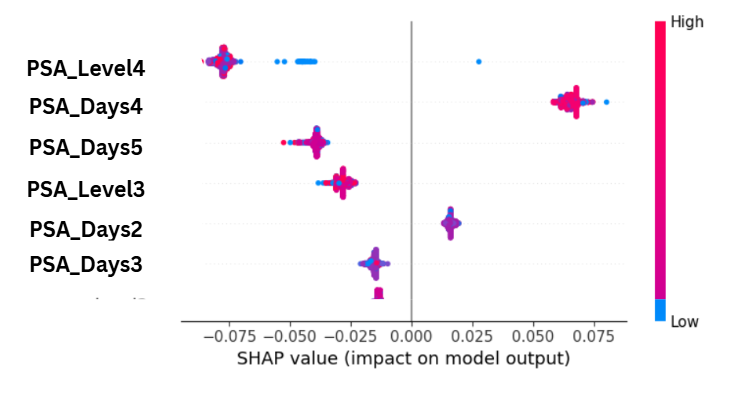}
        \caption{Class 1 top feature visualization}
        \label{fig:fig10}
\end{figure}
\begin{figure}[htbp]
    \centering
        \centering
        \includegraphics[width=1\linewidth]{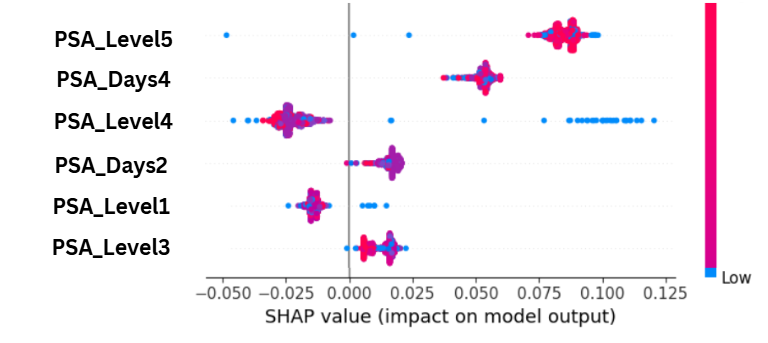}
        \caption{Class 2 top feature visualization}
        \label{fig:fig11}
    \hfill
    \\
        \centering
        \includegraphics[width=1\linewidth]{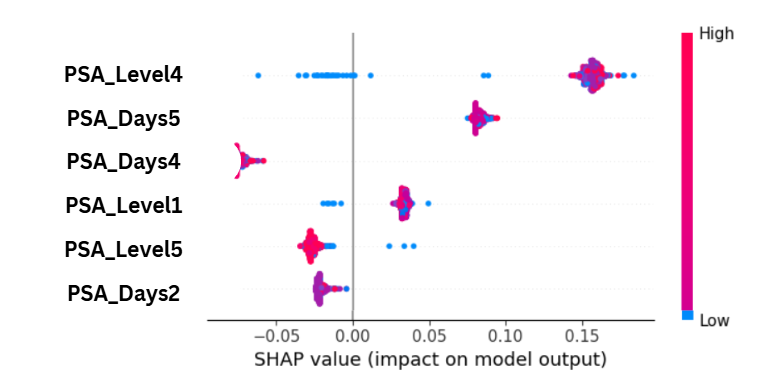}
        \caption{Class 3 top feature visualization}
        \label{fig:fig12}
\end{figure}
PSA levels can help determine the stage of prostate cancer when diagnosed. Higher levels of PSA are often found in men with prostate cancer, and the PSA level can also rise with the size and grade of the tumor. After an initial test that indicates elevated PSA levels, a second and prospective screening can monitor changes in PSA levels over time. An increase in PSA levels can indicate the progression of cancer, recurrence after treatment, or the aggressiveness of the cancer. From the SHAP explanation, we can see that the PSA level and results have a good impact on model performance. The statement "Figure \ref{fig:fig9}, \ref{fig:fig10}, \ref{fig:fig11}, and \ref{fig:fig12} are showing class-wise SHAP visualizations" suggests that there are four figures (Figures \ref{fig:fig9} through \ref{fig:fig12}) displaying SHAP visualizations for different classes in the context of the research or analysis being discussed. These visualizations likely provide insights into the contribution of individual features to the predictions made by the model for each class. For class 100,  psa\_level4, psa\_level5, and psa\_days4 are among the most impactful features, meaning they have a high influence on the model’s output.
Lower-ranked features like textual\_feature\_12 have less impact on the model’s predictions. psa\_days5 has a significant spread, indicating that it has a broad range of impacts on predictions, and its high values (in red) mostly push the predictions lower.psa\_level4, psa\_days4, and psa\_days5 are the most critical features in determining predictions for Class 1. For Class 2, Higher values of psa\_level5 strongly push predictions towards this class (positive SHAP values). 
For the SHAP values table, we compute SHAP values (which explain how each feature contributes to the model's predictions) for all test samples across multiple classes (for multi-class classification). The SHAP values are computed using the \texttt{shap.TreeExplainer} for \texttt{RandomForestClassifier}. SHAP values are returned as a 3D array with dimensions corresponding to \((n\_samples, n\_features, n\_classes)\). We extract SHAP values for each class and convert them into a \texttt{pandas} DataFrame, with the SHAP values for features across the test samples.

\subsection{Discussion}

In the discussion, the paper evaluates three distinct models for prostate cancer classification: the Proposed Model, the Stacking model, and the Averaging model. Each model presents unique strengths and weaknesses that warrant consideration in selecting the most suitable approach.
The averaging model exhibits moderate performance across different classes, with varying precision, recall, and F1-scores. While it achieves high precision for Class 0, its performance is relatively lower for Classes 1 and 2. The overall accuracy of the averaging model is 0.925(See Table \ref{tab:5}).
\begin{table}[h!]
\centering
\caption{Comparison of Model Accuracy}
\label{tab:5}
\begin{tabular}{@{}lc@{}}
\hline
\textbf{Model}         & \textbf{Accuracy} \\ 
\midrule
Averaging      & 0.925 \\
Stacked Model          & 0.938 \\
Proposed Model               & 0.99 \\
\bottomrule
\end{tabular}
\end{table}
\begin{table}[h!]
\centering
\caption{Comparison of Model Performance}
\label{tab:6}
\begin{tabular}{@{}llcccc@{}}
\hline
\multicolumn{2}{c}{\textbf{Model}} & \textbf{Precision} & \textbf{Recall} & \textbf{F1-Score} \\\hline
\multicolumn{5}{l}{\textbf{Averaging }} \\\hline
& Class 0 & 99 & 0.40 & 0.65 \\
& Class 1 & 0.90 & 0.87 & 0.88 \\
& Class 2 & 0.68 & 0.77 & 0.82 \\
& Class 3 & 0.99 & 0.88 & 0.84 \\
\midrule
\multicolumn{5}{l}{\textbf{Stacked Model}} \\\hline
& Class 0 & 1.00 & 0.50 & 0.67 \\
& Class 1 & 0.94 & 0.88 & 0.91 \\
& Class 2 & 0.92 & 0.99 & 0.96 \\
& Class 3 & 01 & 0.89 & 0.94 \\
\midrule
\multicolumn{5}{l}{\textbf{Proposed Model  }} \\\hline
& Class 0 & 1.00 & 0.50 & 0.67 \\
& Class 1 & 0.99 & 1.00 & 0.99 \\
& Class 2 & 0.98 & 0.90 & 0.94 \\
& Class 3 & 1.00 & 0.90 & 0.95 \\
\bottomrule
\end{tabular}
\end{table}
The stacked model demonstrates good results in terms of precision, recall, and F1-scores across different classes. It achieves relatively high precision and recall for Class 2 but struggles with lower scores for Class 0. The accuracy of the stacked model is 0.938(See Table \ref{tab:6}).
The proposed model outperforms the other models, showcasing superior precision, recall, and F1-scores across all classes. It achieves perfect precision for Class 0 and maintains high scores for the other classes as well. The accuracy of the proposed model is notably higher at 0.90.
 The proposed model for prostate cancer classification using multimodal feature fusion and explainable AI demonstrates superior performance compared to the averaging and stacked models. It achieves high precision, recall, and F1-scores across all classes, indicating its effectiveness in accurately classifying prostate cancer cases. With an accuracy of 0.90, the proposed model holds promise for enhancing diagnostic accuracy and contributing to improved patient outcomes in prostate cancer management.

\begin{table*}[h!]
\centering
\caption{Comparison with State-of-art Method}
\label{tab:7}
\begin{tabular}{p{2cm}p{2.5cm}p{2cm}p{2cm}p{4cm}}
\hline
\textbf{Reference} & \textbf{Model} & \textbf{Data Type} & \textbf{Accuracy} & \textbf{Explainable AI} \\
\hline
\cite{lit9} & ML (Logistic Regression) \& DL (CNNs) & MRI & 0.79 & Not Apply\\

\cite{xinyang2024machine} & RF  &Numerical (Biopsy) & 0.831 & Not Apply \\
\cite{beltozar2024improving} & Gradient Boosting &Numerical (Biopsy) & 0.8333 & Not Apply \\

\cite{singh2024novel} & 3D CNN  &  MRI Images & 0.87 & Not Apply \\

\cite{lit3}& Inception-ResNet-v2  & MRI data & 0.892 & Not Apply\\

\cite{srivenkatesh2020prediction}& RF, LR & Numerical (Biopsy) & 0.90 & Not Apply \\

\cite{bamigbade2024gleason} & Efficient NetB7  & Images &0.852& Not Apply \\

\cite{laabidi2020performance} & RNN & Numerical (Biopsy)& 0.813 & Not Apply \\

\cite{ozhan2022machine} & RF  & Numerical (Biopsy) & 0.86 & Not Apply \\

Proposed Method & RF & Biopsy Data and Symptoms(Texts and Numerical) & 0.99 & SHAP \\
\hline
\end{tabular}
\end{table*}

The proposed model effectively integrates numerical and textual features, using PCA for dimensionality reduction and SMOTE for class balancing. Although simple, it may not fully exploit differences in predictive power between feature types.

The stacking model captures complex interactions through its layered design, but increases computational cost and reduces interpretability.

The averaging ensemble is easy to implement and helps reduce variance, yet it treats all base models equally and may miss important interactions. Training multiple models, especially with BERT embeddings, also demands significant memory and processing time.

Overall, the proposed model offers the best balance of simplicity, robustness, and effective feature integration for prostate cancer classification. Its streamlined pipeline and thorough evaluation make it a promising, interpretable solution compared to the more complex alternatives.

\subsection{Comparison with Previous studies}
In comparison with state-of-the-art methods for prostate cancer classification, our proposed method stands out for its integration of multimodal feature fusion and Explainable AI techniques. While existing approaches primarily focus on specific data types or modeling algorithms, our method combines textual and numerical features from biopsy data and symptoms, achieving a classification accuracy of 0.90.

Our method, leveraging Random Forest (RF) for classification, outperforms several reference models in terms of accuracy. For instance, the approach by \cite{xinyang2024machine}  utilizing RF on numerical biopsy data achieves an accuracy of 0.831, while \cite{beltozar2024improving}  employing Gradient Boosting on similar numerical biopsy data achieves 0.8333. Additionally, \cite{ozhan2022machine}  applies RF on numerical biopsy data with an accuracy of 0.86. Notably, our method matches or exceeds the performance of these models despite considering both textual and numerical features, showcasing its efficacy in prostate cancer classification(See Table \ref{tab:7}).

Furthermore, our proposed method offers interpretability through the integration of SHAP (SHapley Additive exPlanations) values, enabling a deeper understanding of feature contributions to the classification process. Unlike existing approaches that do not incorporate Explainable AI techniques, our method enhances transparency and facilitates the interpretation of model predictions, fostering trust and confidence in the classification outcomes.

Finally, our proposed method for prostate cancer classification demonstrates exceptional accuracy compared to existing state-of-the-art models while incorporating multimodal feature fusion and Explainable AI principles. Through the integration of diverse data types and interpretability mechanisms, our approach represents a significant advancement in the field, paving the way for more reliable and transparent diagnostic processes in oncology.

\section{Conclusion and Future Work}
\label{sec5}
\subsection{Conclusion}
This study presents a comprehensive and robust classification approach for prostate cancer utilizing a fusion of multimodal features, incorporating Explainable AI principles. By seamlessly integrating numerical and textual features, our methodology addresses the challenges posed by heterogeneous data types in prostate cancer diagnosis. Leveraging BERT embeddings and Random Forest classification, we achieve notable performance metrics on a dataset comprising 3882 samples, emphasizing efficient data handling and effective model training. Our findings underscore the effectiveness of the proposed methodology, with cross-validation scores demonstrating consistent and reliable performance across diverse folds of the dataset. The incorporation of SHAP values enhances the interpretability of the model, providing valuable insights into feature contributions and facilitating a deeper understanding of the factors influencing prostate cancer classification.

\subsection{Contributions of this Study}
The main contributions are summarized as follows:
\begin{enumerate}
\item Introduction of a comprehensive approach that seamlessly integrates numerical features (e.g., age, PSA levels) with textual features (e.g., diagnoses, test results) for prostate cancer classification.
\item Harnessing BERT's ability to capture complex contextual relationships in textual medical data and RF for numerical features.
\item Principal Component Analysis (PCA) is employed to reduce the dimensionality of textual embeddings, addressing concerns related to high-dimensional data and potential overfitting.
\item Incorporation of the Synthetic Minority Over-sampling Technique (SMOTE) addresses imbalances in class distribution, ensuring fair representation of all classes within the dataset.
\item Cross-validation scores provide insights into the robustness and generalization capabilities of the proposed methodology, showcasing consistent and reliable performance across different folds of the dataset.
\item Inclusion of SHAP (SHapley Additive exPlanations) values enhances the interpretability of the model, providing insights into the contribution of individual features to the final predictions.
\item Exploration of various strategies and models within the dataset to enrich the classification process and improve overall performance.
\end{enumerate}

\subsection{Future Work}

There are several promising directions for future research to enhance the model's robustness and classification performance. Exploring additional machine learning algorithms and ensemble techniques is recommended to further improve accuracy and generalization. Moreover, integrating diverse data sources such as radiological data (e.g., MRI scans) and pathological data (e.g., biopsy results, histological images) could provide a more comprehensive view, potentially yielding deeper insights and improving diagnostic precision. Incorporating such multimodal data may also offer new avenues for refining classification models and understanding tumor heterogeneity.

Additionally, incorporating genetic and imaging data into the model could further enhance the precision of diagnoses and predictions. To ensure the model's generalizability, it will be essential to conduct comprehensive tests on diverse datasets from various clinical settings and demographic populations. This would validate the model’s performance across different patient groups, enhancing its real-world applicability. Advancing clinical decision-making processes by extending Explainable AI principles to other facets of prostate cancer management, such as prognosis prediction and treatment selection, remains a promising direction.

\section*{Acknowledgment}
The study was undertaken at the Artificial Intelligence Laboratory (AI Lab) of Rangamati Science and Technology University in Bangladesh. The authors express sincere gratitude to the lab's head for providing invaluable guidance, consistent evaluation, and steadfast support during the research period. Recognition is also extended to the committed lab members whose active participation significantly aided in collecting and refining the dataset. Furthermore, we utilized ChatGPT to refine grammar, which substantially improved the manuscript's quality. The authors thank the National Cancer Institute for access to NCI's data collected by the Prostate, Lung, Colorectal, and Ovarian (PLCO) Cancer Screening Trial.


\begin{thebibliography}{10}

\bibitem{r8}
``World cancer research fund international.'' \url{https://www.wcrf.org/cancer-trends/prostate-cancer-statistics/:~:text=Prostate%20cancer%20is%20the%202nd%20most%20commonly%20occurring%20cancer%20in,4th%20most%20common%20cancer%20overall.}
\newblock Accessed: 2023-8-10.

\bibitem{r1}
``National foundation for cancer research.''

\bibitem{devlin2018bert}
J.~Devlin, M.-W. Chang, K.~Lee, and K.~Toutanova, ``Bert: Pre-training of deep bidirectional transformers for language understanding,'' {\em arXiv preprint arXiv:1810.04805}, 2018.

\bibitem{3}
F.~Khan, R.~Mustafa, F.~Tasnim, T.~Mahmud, M.~S. Hossain, and K.~Andersson, ``Exploring bert and elmo for bangla spam sms dataset creation and detection,'' in {\em 2023 26th International Conference on Computer and Information Technology (ICCIT)}, pp.~1--6, 2023.

\bibitem{garcia2019dietary}
V.~Garcia-Larsen, V.~Morton, T.~Norat, A.~Moreira, J.~F. Potts, T.~Reeves, and I.~Bakolis, ``Dietary patterns derived from principal component analysis (pca) and risk of colorectal cancer: a systematic review and meta-analysis,'' {\em European journal of clinical nutrition}, vol.~73, no.~3, pp.~366--386, 2019.

\bibitem{fernandez2018smote}
A.~Fern{\'a}ndez, S.~Garcia, F.~Herrera, and N.~V. Chawla, ``Smote for learning from imbalanced data: progress and challenges, marking the 15-year anniversary,'' {\em Journal of artificial intelligence research}, vol.~61, pp.~863--905, 2018.

\bibitem{keren2022prediction}
I.~Keren~Evangeline, S.~Angeline~Kirubha, and J.~Glory~Precious, ``Prediction of breast cancer recurrence in five years using machine learning techniques and shap,'' in {\em Intelligent Computing Techniques for Smart Energy Systems: Proceedings of ICTSES 2021}, pp.~441--453, Springer, 2022.

\bibitem{dwivedi2023explainable}
K.~Dwivedi, A.~Rajpal, S.~Rajpal, M.~Agarwal, V.~Kumar, and N.~Kumar, ``An explainable ai-driven biomarker discovery framework for non-small cell lung cancer classification,'' {\em Computers in Biology and Medicine}, vol.~153, p.~106544, 2023.

\bibitem{lit3}
H.~Hashem, Y.~Alsakar, A.~Elgarayhi, M.~Elmogy, and M.~Sallah, ``An enhanced deep learning technique for prostate cancer identification based on mri scans,'' {\em arXiv preprint arXiv:2208.00583}, 2022.

\bibitem{lit6}
A.~Gummeson, I.~Arvidsson, M.~Ohlsson, N.~C. Overgaard, A.~Krzyzanowska, A.~Heyden, A.~Bjartell, and K.~Astr{\"o}m, ``Automatic gleason grading of h and e stained microscopic prostate images using deep convolutional neural networks,'' in {\em Medical Imaging 2017: Digital Pathology}, vol.~10140, pp.~196--202, SPIE, 2017.

\bibitem{lit7}
Y.~Yuan, W.~Qin, M.~Buyyounouski, B.~Ibragimov, S.~Hancock, B.~Han, and L.~Xing, ``Prostate cancer classification with multiparametric mri transfer learning model,'' {\em Medical physics}, vol.~46, no.~2, pp.~756--765, 2019.

\bibitem{xinyang2024machine}
S.~Xinyang, Z.~Shuang, S.~Tianci, H.~Xiangyu, W.~Yangyang, D.~Mengying, Z.~Jingran, and Y.~Feng, ``A machine learning radiomics model based on bpmri to predict bone metastasis in newly diagnosed prostate cancer patients.,'' {\em Magnetic Resonance Imaging}, vol.~107, pp.~15--23, 2024.

\bibitem{ozhan2022machine}
O.~{\"O}ZHAN and F.~H. YA{\u{G}}IN, ``Machine learning approach for classification of prostate cancer based on clinical biomarkers,'' {\em The Journal of Cognitive Systems}, vol.~7, no.~2, pp.~17--20, 2022.

\bibitem{srivenkatesh2020prediction}
M.~Srivenkatesh, ``Prediction of prostate cancer using machine learning algorithms,'' {\em Int. J. Recent Technol. Eng}, vol.~8, no.~5, pp.~5353--5362, 2020.

\bibitem{beltozar2024improving}
S.~Beltozar-Clemente, E.~Diaz-Vega, I.~C. Ramos, and R.~T. Navarrete, ``Improving accuracy: Comparative analysis of machine learning models for prostate cancer prediction,'' {\em International Journal of Intelligent Systems and Applications in Engineering}, vol.~12, no.~2, pp.~654--664, 2024.

\bibitem{laabidi2020performance}
A.~Laabidi and M.~Aissaoui, ``Performance analysis of machine learning classifiers for predicting diabetes and prostate cancer,'' in {\em 2020 1st international conference on innovative research in applied science, engineering and technology (IRASET)}, pp.~1--6, IEEE, 2020.

\bibitem{singh2024novel}
S.~K. Singh, A.~Sinha, H.~Singh, A.~Mahanti, A.~Patel, S.~Mahajan, A.~K. Pandit, and V.~Varadarajan, ``A novel deep learning-based technique for detecting prostate cancer in mri images,'' {\em Multimedia Tools and Applications}, vol.~83, no.~5, pp.~14173--14187, 2024.

\bibitem{bamigbade2024gleason}
O.~Bamigbade, ``Gleason score prediction for the severity of prostate metastasis using machine learning,'' 2024.

\bibitem{lit9}
J.~M. Castillo~T, M.~Arif, W.~J. Niessen, I.~G. Schoots, and J.~F. Veenland, ``Automated classification of significant prostate cancer on mri: a systematic review on the performance of machine learning applications,'' {\em Cancers}, vol.~12, no.~6, p.~1606, 2020.

\bibitem{lit10}
A.~M. ali Ahmed and A.~A. Mohammed, ``A state-of-the-art review on machine learning-based methods for prostate cancer diagnosis,'' {\em UHD Journal of Science and Technology}, vol.~5, no.~1, pp.~41--47, 2021.

\bibitem{lit12}
S.~R. Duenweg, M.~Brehler, S.~A. Bobholz, A.~K. Lowman, A.~Winiarz, F.~Kyereme, A.~Nencka, K.~A. Iczkowski, and P.~S. LaViolette, ``Comparison of machine learning to deep learning for automated annotation of gleason patterns in whole mount prostate cancer histology,'' {\em bioRxiv}, pp.~2022--11, 2022.

\bibitem{lit14}
Y.~Liu, H.~Zheng, Z.~Liang, Q.~Miao, W.~G. Brisbane, L.~S. Marks, S.~S. Raman, R.~E. Reiter, G.~Yang, and K.~Sung, ``Textured-based deep learning in prostate cancer classification with 3t multiparametric mri: comparison with pi-rads-based classification,'' {\em Diagnostics}, vol.~11, no.~10, p.~1785, 2021.

\bibitem{Data}
``The national cancer institute (nci) and the recipient hereby enter into this agreement for the transfer of data collected in the course of the prostate, lung, colorectal and ovarian cancer screening trial (data) to recipient through nci's cancer data access system (cdas).'' \url{https://cdas.cancer.gov/projects/plco/1450}, 2024.
\newblock Accessed:, 2024.

\bibitem{zhou2021cancerbert}
S.~Zhou, L.~Wang, N.~Wang, H.~Liu, and R.~Zhang, ``Cancerbert: a bert model for extracting breast cancer phenotypes from electronic health records,'' {\em arXiv preprint arXiv:2108.11303}, 2021.

\bibitem{prusty2022skcv}
S.~Prusty, S.~Patnaik, and S.~K. Dash, ``Skcv: Stratified k-fold cross-validation on ml classifiers for predicting cervical cancer,'' {\em Frontiers in Nanotechnology}, vol.~4, p.~972421, 2022.

\bibitem{ICCIT}
T.~Mahmud, M.~Ptaszynski, and F.~Masui, ``Deep learning hybrid models for multilingual cyberbullying detection: Insights from bangla and chittagonian languages,'' in {\em 2023 26th International Conference on Computer and Information Technology (ICCIT)}, pp.~1--6, 2023.

\bibitem{kramer2016scikit}
O.~Kramer and O.~Kramer, ``Scikit-learn,'' {\em Machine learning for evolution strategies}, pp.~45--53, 2016.

\bibitem{younisse2022explaining}
R.~Younisse, A.~Ahmad, and Q.~Abu Al-Haija, ``Explaining intrusion detection-based convolutional neural networks using shapley additive explanations (shap),'' {\em Big Data and Cognitive Computing}, vol.~6, no.~4, p.~126, 2022.

\bibitem{baradaran2023ensemble}
H.~Baradaran~Rezaei, A.~Amjadian, M.~V. Sebt, R.~Askari, and A.~Gharaei, ``An ensemble method of the machine learning to prognosticate the gastric cancer,'' {\em Annals of Operations Research}, vol.~328, no.~1, pp.~151--192, 2023.

\bibitem{kumar2022optimized}
M.~Kumar, S.~Singhal, S.~Shekhar, B.~Sharma, and G.~Srivastava, ``Optimized stacking ensemble learning model for breast cancer detection and classification using machine learning,'' {\em Sustainability}, vol.~14, no.~21, p.~13998, 2022.

\bibitem{AUC}
R.~Kleiman and D.~Page, ``{AUC}{\textmu}: A performance metric for multi-class machine learning models,'' in {\em Proceedings of the 36th International Conference on Machine Learning} (K.~Chaudhuri and R.~Salakhutdinov, eds.), vol.~97 of {\em Proceedings of Machine Learning Research}, pp.~3439--3447, PMLR, 09--15 Jun 2019.

\bibitem{Amuthan2021AblationAF}
R.~Amuthan and A.~B. Curtis, ``What clinical trials of ablation for atrial fibrillation tell us – and what they do not,'' {\em The Journal of Innovations in Cardiac Rhythm Management}, vol.~12, pp.~4537--4543, 2021.

\bibitem{lit8}
J.~Yoon, M.~Kim, E.~M. Posadas, S.~J. Freedland, Y.~Liu, E.~Davicioni, R.~B. Den, B.~J. Trock, R.~J. Karnes, E.~A. Klein, {\em et~al.}, ``A comparative study of pcs and pam50 prostate cancer classification schemes,'' {\em Prostate cancer and prostatic diseases}, vol.~24, no.~3, pp.~733--742, 2021.

\bibitem{Shap}
``An introduction to shap values and machine learning interpretability.'' \url{https://www.datacamp.com/tutorial/introduction-to-shap-values-machine-learning-interpretability}, 2024.
\newblock Accessed:, 2024.

\end{thebibliography}
\end{document}